\documentclass[preprint,12pt]{elsarticle}

\usepackage{amssymb}
\usepackage{amsmath}

\usepackage{graphicx}
\usepackage{amsthm}
\usepackage{booktabs}
\usepackage{algorithm}
\usepackage{algorithmic}
\usepackage[switch]{lineno}
\usepackage{multirow}
\usepackage{xcolor} 

\usepackage{balance}

\usepackage[normalem]{ulem}
\usepackage{makecell}
\usepackage[most]{tcolorbox}
\usepackage{lmodern}

\makeatletter
\def\thefnote{%
  \ifcase\c@fnote\or
    $\dagger$\or
    $\ddagger$\or
    $\mathsection$\or
    $\mathparagraph$%
  \else
    \@ctrerr
  \fi}
\makeatother

\journal{Neural Networks}

\begin{document}

\begin{frontmatter}



\title{HiFiVe: High-Fidelity Vehicle Generation Leveraging Auto-Regressive 2D Generative Priors}




\author[1,2,3]{Hongli Xiao\fnref{equal}}
\ead{honglixiao@sjtu.edu.cn}

\author[4]{Youjian Zhang\fnref{equal}}
\ead{Youjian.Zhang@cn.bosch.com}

\author[5]{Qi Zheng}
\ead{qiz@szu.edu.cn}

\author[3]{Zhaohui Hu}
\ead{hzhhx@nudt.edu.cn}


\author[1]{Yaohui Jin}
\ead{jinyh@sjtu.edu.cn}

\author[2]{Xiaoguang Ren}
\ead{rxg_nudt@126.com}

\author[3]{Wenjing Yang}
\ead{wenjing.yang@nudt.edu.cn}

\author[3]{Long Lan\corref{cor1}}
\ead{long.lan@nudt.edu.cn}

\fntext[equal]{These authors contributed equally to this work.}
\cortext[cor1]{Corresponding author.} 
\nonumnote{This work was supported by the National Natural Science Foundation of China (No. 62376282) and the Science and Technology Innovation Program of Hunan Province (No. 2025RC3117).}

\affiliation[1]{organization={MoE Key Lab of Artificial Intelligence, AI Institute, Shanghai Jiao Tong University},
            city={Shanghai}, 
            postcode={200240}, 
            country={China}}

\affiliation[2]{
            organization={Academy of Military Science},
            city={Beijing},
            postcode={100071}, 
            country={China}}

\affiliation[3]{
  organization={College of Computer Science and Technology, National University of Defense Technology},
  city={Changsha},
  postcode={410073},
  country={China}
}

\affiliation[4]{
  organization={Bosch Innovation Software Development (Wuxi) Co., Ltd.},
  city={Wuxi},
  postcode={214028},
  country={China}
}

\affiliation[5]{
  organization={Shenzhen University},
  city={Shenzhen},
  postcode={518060},
  country={China}
}



\begin{abstract}
    Existing 3D vehicle  generation methods often suffer from low geometric fidelity and blurry textures, hindering their downstream applications.
    While recent works adopt multi-view diffusion  models for high-fidelity texture, they are often constrained by fixed viewpoints, limited resolution, and a reliance on costly fine-tuning to achieve cross-view consistency.
    In this paper, we propose \textbf{HiFiVe}, a training-free framework for high-fidelity vehicle modeling through joint texture and geometry enhancement by imposing 3D geometric constraints to anchor 2D generative priors.
    Specifically, we propose an auto-regressive texture refinement pipeline that progressively synthesizes high-resolution textures from arbitrary viewpoints. To ensure cross-view consistency, the coarse geometry serves as a synchronization prior, conditioning each generation step on previously synthesized frames via depth-based warping and multi-view texture fusion. Moreover, the inherent symmetry of vehicles is exploited to mitigate error accumulation.
    Finally, high-frequency surface details are recovered by refining the mesh geometry using normal maps estimated from the enhanced textures.
    Extensive experiments on synthetic and real-world vehicle datasets demonstrate that our method  significantly improves both geometric detail and texture quality compared to state-of-the-art baselines.
    Project page: https://honglixiao.github.io/hifive.github.io/.
\end{abstract}



\begin{keyword}
3D Generation \sep Auto-Regressive Generation \sep 2D Generative Priors




\end{keyword}

\end{frontmatter}


\section{Introduction}

High-quality 3D vehicle assets are a critical component for a wide range of applications, including autonomous driving simulation \cite{liu2025protocar,yang2025genassets}, synthetic data generation \cite{li2024drivingdiffusion,yang2023unisim}, game engine content creation \cite{zhao2025hunyuan3d2.0,lai2025hunyuan3d2.5}, and virtual and augmented reality \cite{kerbl20233dgs,tang2024lgm}. 
Traditionally, realistic 3D vehicle models with accurate geometry and detailed textures are manually crafted by professionals using specialized tools, which is time-consuming and labor-intensive.

Driven by the success of diffusion models~\cite{ho2020denoising,rombach2022high}, 3D generation \cite{pooledreamfusion,long2024wonder3d,xiang2024trellis,zhao2025hunyuan3d2.0} has developed rapidly in recent years. 
From 2D Lifting via 2D diffusion prior \cite{pooledreamfusion,chen2023fantasia3d} to native 3D generation~\cite{xiang2024trellis,he2025sparseflex}, 3D generative models have achieved remarkable progress in geometric precision and texture fidelity. 
However, as these models are typically trained on synthetic 3D datasets~\cite{deitke2023objaverse,deitke2023objaverse_xl}, they often struggle to generalize to in-the-wild scenarios. Consequently, they fail to recover fine-grained geometry and realistic textures when encountering the complexity of real-world data.

In contrast, 2D generative models \cite{esser2024SD3.5,flux2024} leverage real-world training data (orders of magnitude larger than existing 3D datasets) to capture richer visual priors for photorealistic, high-textured synthesis. Motivated by this, recent works have sought to leverage these potent 2D priors for high-fidelity 3D texture generation. For example, \cite{zhao2025hunyuan3d2.0,yang2025wonder3d++} adopt multi-view diffusion models for texture generation and refinement, where images from multiple angles are synthesized jointly in a single diffusion process.
To maintain cross-view consistency, these methods typically rely on fine-tuning on curated 3D datasets to obtain multi-view renderings, which makes them suffer from several drawbacks: i) They are constrained by predefined view numbers and camera layouts, which limits the viewpoint coverage. ii) Simultaneously generating multiple views incurs significant memory costs, forcing a reduction in per-view image quality and resolution. In some methods \cite{wu2024unique3d,yang2025wonder3d++}, a super-resolution module is still required. iii) The cumbersome fine-tuning process prevents these methods from immediately leveraging emerging SOTA 2D diffusion models without retraining.


To address these limitations, we propose HiFiVe, a training-free framework for joint texture and geometry enhancement. 
The core idea is to leverage powerful pretrained 2D generative priors for synthesizing high-frequency details, while explicitly enforcing 3D geometric constraints as a synchronization mechanism to maintain cross-view consistency. 
For texture refinement, we propose an autoregressive refinement pipeline that sequentially synthesizes photorealistic images and projects them back onto the mesh surface. 
Specifically, we adopt Flux.2~\cite{flux-2-2025} as our generative backbone, leveraging its superior image synthesis and editing capabilities. Starting from an initial rendering of the low-quality mesh, Flux.2 first refines it into a high-fidelity image that serves as the anchor for our autoregressive pipeline. For subsequent steps, all previously refined frames are warped to the current perspective as conditioning signal via depth-based projection. To resolve inter-view conflicts during multi-view warping, we introduce a multi-view texture fusion strategy that assigns higher confidence to source pixels whose viewing directions are better aligned with the  surface normal. The generator (Flux.2) then processes the resulting incomplete imagery, infilling missing regions while strictly preserving the warped content. This entire process is synchronized through depth maps and out-painting masks rendered from the coarse geometry, ensuring global cross-view consistency without the need for training additional transformer blocks.
Unlike the multi-view generation mechanism, our autoregressive approach offers the flexibility to synthesize an arbitrary number of views at significantly higher resolutions.

To prevent error accumulation inherent in long autoregressive chains, we leverage the structural symmetry of vehicles to optimize our generation sequence. By carefully routing the synthesis path, we ensure that all generation viewpoints remain within three steps of the initial anchor. This constraint effectively suppresses error propagation and maintains consistent texture quality across previously unseen regions.


Finally, high-frequency surface details are recovered by refining the mesh geometry using normal maps estimated from the enhanced textures. 
{
Since the normals predicted from texture images mainly contribute high-frequency geometric details, we introduce a frequency-adaptive weighting strategy that explicitly modulates the contribution of different frequency components during the optimization process.
Specifically, we assign larger weights to high-frequency residuals of the rendered normals, thereby encouraging the restoration of sharp and detailed surface structures while maintaining overall geometric stability.
As a result, HiFiVe produces high-quality vehicle models with both  photorealistic textures and detailed geometry.
}


Our main contributions are summarized as follows:
\begin{itemize}
    \item 
    We propose HiFiVe, a training-free framework for the joint enhancement of texture and geometry in low-quality vehicle meshes by anchoring powerful 2D generative priors to 3D geometric constraints. 

    \item 
    We introduce a geometry-synchronized autoregressive pipeline, that iteratively synthesizes high-resolution textures from arbitrary viewpoints. To ensure global consistency, we synchronize the generation process through depth-based warping and multi-view texture fusion. Furthermore, we recover intricate geometry details with high-frequency normals estimated from the refined texture.
    

    \item Extensive experiments on both synthetic and real-world vehicle datasets demonstrate that our method significantly outperforms state-of-the-art methods in both visual fidelity and geometric details.

\end{itemize}





\section{Related Works}

\subsection{General 3D Generation}

In recent years, 3D generation has drawn significant attention as advances in diffusion models~\cite{rombach2022high, shen2023contrastive} and large-scale 3D datasets~\cite{deitke2023objaverse,deitke2023objaverse_xl, shen2024graph} have enabled the creation of high-fidelity 3D models from 2D images. 
Earlier methods~\cite{pooledreamfusion,lin2023magic3d,qianmagic123,chen2023fantasia3d,zhou2023sparsefusion, shen2022unsupervised, shen2024multiple} typically follow the paradigm of per-shape optimization by distilling priors from 2D generative models to assist 3D generation tasks. 
Zero123~\cite{liu2023zero} has revealed the potential of fine-tuning the pre-trained 2D Stable Diffusion model for novel view synthesis but suffers from severe view-inconsistency issues since it generates each novel view independently.
Subsequent works proposed multi-view diffusion models~\cite{shi2023MVDream,tang2024mvdiffusion++,wang2023imagedream} that can generate multiple images simultaneously. 
Nevertheless, these models usually perform well on global semantic consistency and struggle to achieve local geometry consistency. 
Furthermore, they can only generate a fixed number of images with predefined viewpoints, posing challenges to subsequent 3D reconstruction.
Meanwhile, native 3D generation (e.g., TRELLIS~\cite{xiang2024trellis}) is another representative paradigm that learns the distribution of 3D representations from large-scale 3D datasets and achieves remarkable structural consistency than 2D lifting methods.
However, due to the scarcity of high-quality training data, these native 3D models often yield coarse geometry and blurry or overly smoothed textures, especially for realistic vehicle assets.



\subsection{ Car-specific 3D Generation }

To facilitate the development of autonomous driving and vehicle modeling, several dedicated methods have emerged for 3D car generation. 
GINA-3D~\cite{shen2023gina} introduces a transformer-based generative model that uses driving data from both the camera and LiDAR to create 3D implicit neural assets. 
DreamCar~\cite{du2024dreamcar} proposes a four-stage 3D car reconstruction method leveraging car-specific prior given one to five supervision images. Its performance significantly degrades when only single-view inputs are available.
Car-Studio~\cite{liu2024car} utilizes an encoder-decoder based network to learn priors and regresses NeRF representations from single car images. 
However, these works suffer from high training and rendering costs and often exhibit artifacts upon large view changes due to overfitting.
Drive-1-to-3~\cite{lin2024drive} constructs a vehicle-specific multi-view diffusion model fine-tuned from Free3D~\cite{zheng2024free3d}. 
RGM~\cite{chen2024rgm} proposes a relightable 3D-GS generative model for generating 3D car assets with material properties. Nevertheless, it heavily relies on a self-curated dataset, which may limit its generalization to real-world cars.

\begin{figure*}[!ht]
\begin{center}
\includegraphics[width=\linewidth]{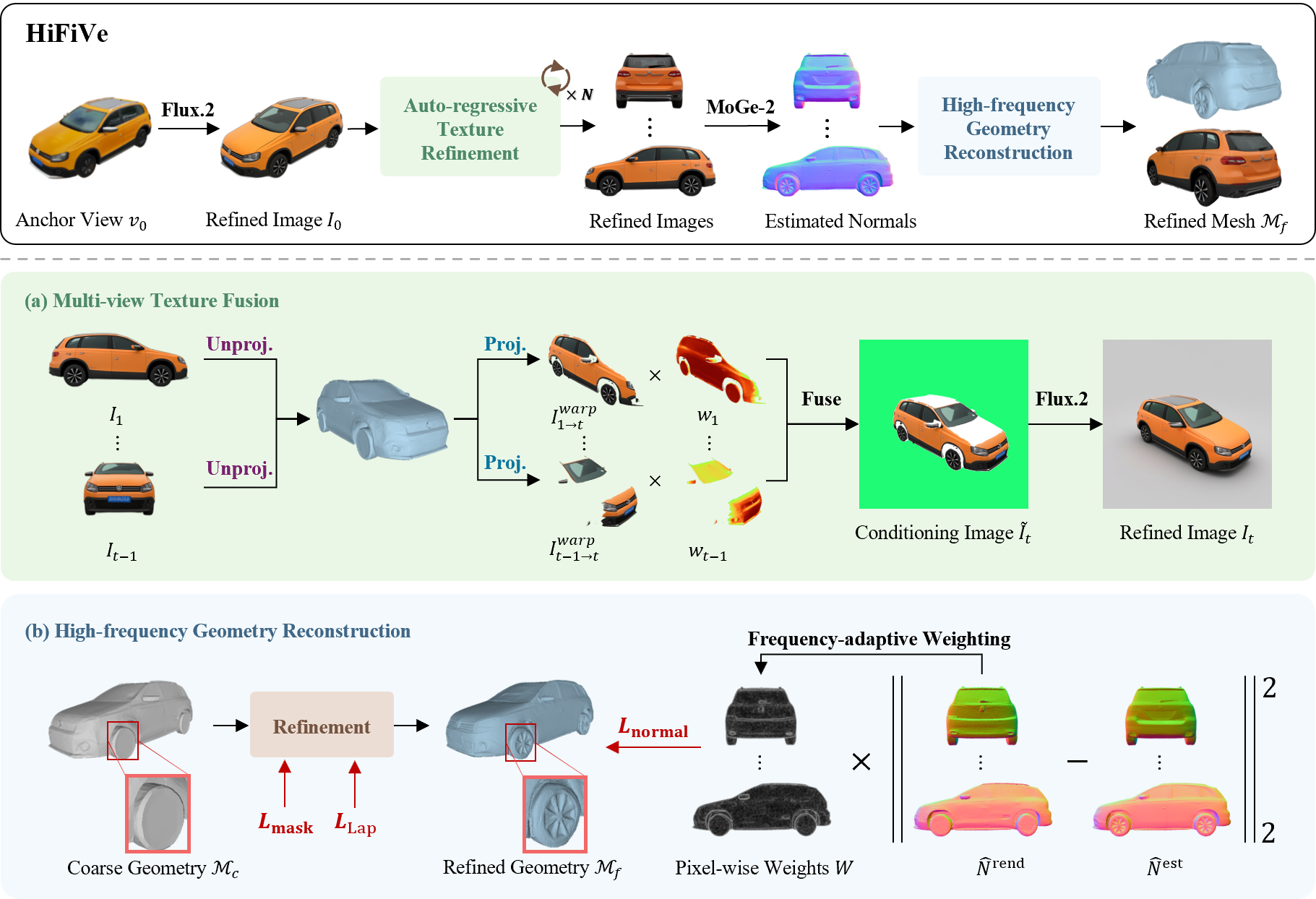}
\end{center}
\vspace{-6mm}
   \caption{ 
    Pipeline of HiFiVe.
    Starting from a coarse vehicle mesh, the framework generates multi-view texture images via geometry-synchronized autoregressive refinement with confidence-aware fusion.
    Then normal maps estimated from the refined textures guide high-frequency geometry refinement with frequency-adaptive weighting, resulting in a high-quality 3D vehicle model.
       }
\label{fig:pipeline}
\end{figure*}

\section{Methods}

In this section, we present the proposed HiFiVe framework. 
Given a coarse 3D mesh $\mathcal{M}_{c}$ (e.g. generated by TRELLIS~\cite{xiang2024trellis}), our goal is to generate a high-fidelity textured mesh $\mathcal{M}_{f}$. 
As illustrated in Fig.~\ref{fig:pipeline}, HiFiVe consists of three main components: 
(i) an autoregressive texture refinement pipeline that progressively synthesizes high-resolution multi-view textures under explicit geometric synchronization, 
(ii) a vehicle-specific symmetric generation scheme to mitigate error accumulation during long-term autoregressive synthesis, and 
(iii) a geometry refinement stage that recovers high-frequency surface details using normals estimated from the enhanced textures.


\subsection{Autoregressive Texture Refinement}
\label{sec:autoregressive-texture}

To distill high-fidelity details from 2D generative priors while maintaining 3D consistency, we propose an autoregressive texture refinement pipeline under explicit geometric constraints. 
Instead of generating multiple views simultaneously with limited resolution, we synthesize a sequence of views iteratively. 
At each step $t$, the generation of the current view $v_t$ is conditioned on the history of all previously refined views $\mathcal{H}_{t-1} = \{v_0, ..., v_{t-1}\}$ to ensure texture coherence.


\subsubsection{3D-Conditioned Texture Propagation}


The core challenge of autoregressive generation lies in how to effectively maintain the cross-view consistency from the history $\mathcal{H}_{t-1}$ to the current target view $v_t$. 
We treat the coarse mesh $\mathcal{M}_{c}$ as a geometric synchronization prior to guide the generation process.

Specifically, for each viewpoint $v_i$ with camera pose $P_i$, we first extract its geometric attributes from $\mathcal{M}_c$ via the rasterization operator $\mathcal{R}$:
\begin{equation}
    (D_i, O_i) = \mathcal{R}(\mathcal{M}_c, P_i),
\end{equation}
where $D_i \in \mathbb{R}^{H \times W}$ denotes the depth map and $O_i \in \{0, 1\}^{H \times W}$ is the binary occupancy mask indicating valid vehicle regions projected onto the current view.

Suppose we have a source view $v_i \in \mathcal{H}_{t-1}$ with the refined image $I_i$ and rendered depth map $D_i$, we propagate its texture to the target view via a depth-based warping operator.
The warping process is formulated as:
\begin{equation}
    I^{warp}_{i\to t}
    \;=\;
    \mathcal{W}_{i\to t}\!\left(I_i;\, D_i,\, D_t,\, P_i,\, P_t\right),
    \label{eq:depth_warp}
\end{equation}
where $\mathcal{W}(\cdot)$ denotes the warping operator parameterized by the camera poses $P$ and depth maps $D$.
This operation transfers the visible pixels from the source to the target, yielding the warped image $I^{warp}_{i\to t}$ indicating successfully projected pixels. This warped image will then serve as the conditioning pixels for the subsequent generation on target view.


 \subsubsection{Multi-view Texture Fusion}
 \label{sec:multiview_fusion}

As autoregression continues, the history $\mathcal{H}_{t-1}$ accumulates multiple refined views. 
Relying on a single preceding view to condition the current generation is suboptimal, instead, we want to integrate texture information from all previously visited viewpoints.
However, naively averaging warped pixels from different views is prone to blurring and ghosting artifacts due to inevitable misalignments. 
Inspired by the normal-weighted blending scheme used in mesh coloring~\cite{wu2024unique3d,yang2025wonder3d++}, we conduct a multi-view texture fusion method by assigning higher confidence to source pixels  whose view directions align closely with the corresponding surface normals. The underlying intuition is that observations from near-orthogonal viewpoints are inherently more reliable and provide the most faithful color information.




Specifically, each pixel $u$ in the target view $v_t$ may receive reprojected texture information from multiple source views $I^{\text{warp}}_{i \to t}(u)$. 
To prioritize the most reliable observations, we calculate the contribution weight $w_i(u)$ based on the angular alignment between the view-space normal $\mathbf{n_i}(u)$  from coarse mesh  and the viewing direction $\mathbf{d}_i(u)$ from the source view :
\begin{equation}
    w_{i}(u) = 
    \begin{cases} 
    (|\text{cos}({\mathbf{n}_i(u), \mathbf{d}_i(u))|)^2,} 
    & \text{if } |\text{cos}(\mathbf{n}_i(u), \mathbf{d}_i(u)) | > \tau \\
    0, & \text{otherwise}
    \end{cases}
\end{equation}
where { $\text{cos}(\cdot, \cdot)$ } calculates the cosine of the angle between the surface normal and the source view direction,  $\tau$ is a confidence threshold to remove low-confidence reprojected pixels.

Using these confidence weights, we aggregate all warped views into a single fused canvas via weighted averaging:
\begin{equation}
    I^{\text{fused}}_t(u) =
    \frac{\sum_{i=1}^{t-1} w_i(u) \cdot {I^{\text{warp}}_{i \to t}(u)}}{\sum_{i=1}^{t-1} w_i(u) + \epsilon},
\end{equation}
where $\epsilon$ is a small constant for numerical stability. 
This fusion mechanism effectively aggregates the best available texture details from the entire generation history.

Then we convert $I^{fused}_t$ into a conditioning image for the 2D generative model.  Specifically, regions outside the vehicle occupancy mask $O_t$ are assigned a distinct, non-natural background color. Within the mask, validly reprojected pixels from $I^{\text{fused}}_t$ are retained to serve as a geometric-consistent constraint, while the remaining unseen regions are filled with a uniform white value, indicating the targets for generative inpainting (as shown in the example of Fig. \ref{fig:pipeline}).

Finally, the refined image $I_t$ is synthesized by conditioning the 2D generator $\mathcal{G}(\cdot)$ on the composite image $\tilde{I}_t$  and a text prompt $\mathbf{p}$ that instructs the model to preserve all conditional regions and complete only the missing vehicle parts:
\begin{equation}
    I_t = \mathcal{G}(\tilde{I}_t, \mathbf{p}).
\end{equation}



By explicitly marking missing regions with white within the object silhouette, we guide the model to generate coherent textures for the unseen parts while faithfully preserving the visual details propagated from previous views. Thus, this geometry-conditioned autoregressive mechanism ensures the multi-view consistency in a training-free manner.




\begin{figure}[!ht]
\begin{center}
\includegraphics[width=0.65\linewidth]{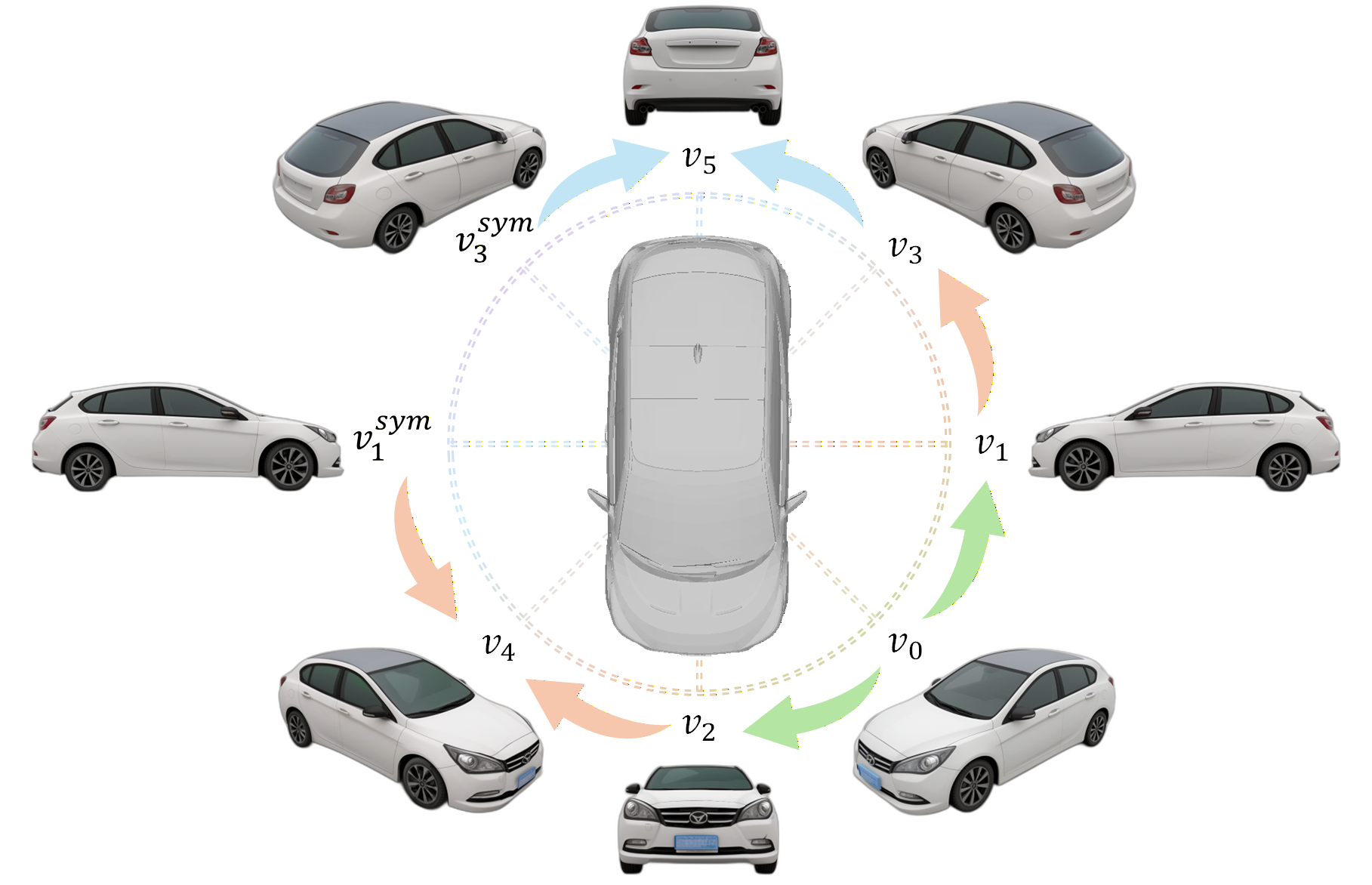}
\end{center}
\vspace{-6mm}
   \caption{  
   An illustrative example of the vehicle-specific generation trajectory.}
\label{fig:trajectory-diagram}
\end{figure}

\subsection{Vehicle-specific Generation Trajectory}
\label{sec:vehicle-specific-generation-trajectory}



Long-term autoregressive generation tends to produce accumulation error, especially when the target generation view locates on the opposite side of the vehicle relative to the initial anchor. Exploiting the inherent symmetry of vehicles, we propose a symmetric generation path that minimizes the autoregressive distance between target views and the initial anchor.


Instead of generating in a naive circular trajectory, we construct a Breadth-First Search (BFS)-style, layer-wise generation scheme.
Specifically, we parameterize each viewpoint $v$ by its azimuth $\theta \in [0^\circ, 360^\circ)$ and elevation angle $\phi$. Starting from an anchor view $v_{0}$, the generation traverses simultaneously along the azimuth $\theta$ in both clockwise and counterclockwise directions. Within each generation round, we process all viewpoints that share the same topological distance from $v_{0}$. Furthermore, we exploit the vehicle's bilateral symmetry by assigning a viewpoint and its mirrored counterpart to the same generation round. The whole generation scheme is illustrated in Fig.~\ref{fig:trajectory-diagram}. This strategy effectively shortens the propagation depth.


Specifically, we define the symmetric counterpart view of a given view $v = (\theta, \phi)$ as:
\begin{equation}
v^{sym} = ((360^\circ - \theta) \bmod 360^\circ, \phi).
\end{equation}
{
We directly obtain the image of symmetric view by horizontally flipping the original image. These mirrored frames are then added to the conditioning history  $\mathcal{H}_{t-1}$ for the subsequent generation round.
}

Overall, this vehicle-specific symmetry prior not only decreases the inference cost of symmetrical views,
but also effectively shortens the distance of autoregression chains, mitigating the error accumulation.

\subsection{High-frequency Geometry Refinement}
\label{sec:geometry}

While the autoregressive texture refinement pipeline produces a set of high-fidelity and view-consistent texture images, the underlying mesh $\mathcal{M}_c$ remains geometrically coarse. 
To restore fine surface details, we further refine the mesh geometry via normal-guided optimization~\cite{Laine2020diffrast,wu2024unique3d}.


Firstly, we apply a state-of-the-art monocular normal estimator~\cite{wang2025moge} to predict the surface normal map $N_i^{est}$ for each refined texture image $I_i$. 
Leveraging the fine details from the enhanced texture, $N_i^{est}$ can effectively recover high-frequency normals, which are missing from the initial coarse mesh. 

Then, we adopt the ISOMER \cite{wu2024unique3d} algorithm to perform mesh optimization guided by multi-view normals.
{ Its coarse-to-fine mesh optimization strategy allows us to initialize the geometry from a coarse mesh and progressively recover local details while maintaining the stability of the global structure.}
To mitigate inconsistencies between monocularly estimated normal maps $N_i^{est}$ across views, we follow its ExplicitTarget (ET) strategy to aggregate the observations on the mesh surface and render them back to each view, producing $\hat{N}_i^{est}$ for supervision.
{This helps reduce conflicting normal supervision and improves the stability of geometry refinement.}
 

To further preserve high-frequency surface details, we introduce a frequency-adaptive weighting strategy on normal loss, which is specifically designed to emphasize high-frequency features during the optimization process.
At each optimization iteration, the current geometry is differentiably rendered to obtain view-dependent normal maps, which are then compared against the pseudo ground-truth normals $\hat{N}_i^{est}$.  
For each rendered normal map $\hat{N}_i^{\mathrm{rend}}$, we compute the frequency-adaptive weight as the residual between the rendered normal map and its low-frequency component:
\begin{equation}
W_i = \mathrm{Norm}\!\left(
\left\| \hat{N}_i^{\mathrm{rend}} - G_{\sigma}\!\left(\hat{N}_i^{\mathrm{rend}}\right) \right\|_2
\right) \cdot M_i,
\end{equation}
{where $G_\sigma(\cdot)$ is the Gaussian blur kernel, and $\mathrm{Norm}$ operator represents a min-max normalization that rescales the residual magnitudes to a $[0, 1]$ range. $M_i$ denotes an eroded occupancy mask $O_i$, which is applied to exclude the edges of vehicle boundary.}
The frequency-weighted normal loss is then derived as:
\begin{equation}
\mathcal{L}_{\mathrm{normal}}
=
\sum_{i}
\sum_{u}
W_i(u)\,
\left\|
\hat{N}^{\mathrm{rend}}_i(u) - \hat{N}^{\mathrm{est}}_i(u)
\right\|_2^2,
\label{eq:loss_geo}
\end{equation}
where $u$ denotes the pixel location and $i$ denotes the view index.

In each optimization step,
the mesh geometry is updated by minimizing the total loss function:
\begin{equation}
\mathcal{L}_{\mathrm{total}}
=
\lambda_{\mathrm{normal}}\, \mathcal{L}_{\mathrm{normal}}
+
\lambda_{\mathrm{mask}}\, \mathcal{L}_{\mathrm{mask}}
+
\lambda_{\mathrm{Lap}}\, \mathcal{L}_{\mathrm{Lap}},
\label{eq:loss_isomer}
\end{equation}
where $\lambda_{\mathrm{normal}}$, $\lambda_{\mathrm{mask}}$, and $\lambda_{\mathrm{Lap}}$ are balancing coefficients.
The mask loss $\mathcal{L}_{\mathrm{mask}}$ is defined as an $\ell_2$ loss between the rendered occupancy mask and the predicted foreground mask to enforce silhouette alignment.
$\mathcal{L}_{\mathrm{Lap}}$ denotes a Laplacian regularization term to promote local surface smoothness.


\section{Experiments}

\subsection{Experimental Setting}

\noindent \textbf{Implementation details.}   
For coarse mesh generation, we choose open-sourced SOTA TRELLIS~\cite{xiang2024trellis} as our baseline.
For texture refinement, we employ the pre-trained Flux.2~\cite{flux-2-2025} model as the 2D generative prior.
Unless otherwise specified, we generate 8 views for each vehicle instance at a resolution of $1024 \times 1024$ with $20$ denoising steps and a classifier-free guidance scale of $4.0$. 
The camera intrinsics remain constant across views, and the field of view is set to $50^\circ$.
For camera extrinsics, the anchor view is initialized with an azimuth angle of $45^\circ$ and an elevation angle of $20^\circ$ for a better coverage of the vehicle.
Subsequent views are generated by uniformly sampling the azimuth at intervals of $45^\circ$, while alternately assigning elevation angles of $0^\circ$ and $20^\circ$.
In the texture fusion stage, the confidence threshold is set to $\tau=0.15$. 
For geometry refinement, we employ MoGe-2~\cite{wang2025moge} for normal estimation.
The mesh geometry is then optimized using the loss defined in Eq.(\ref{eq:loss_isomer}), 
 where $\lambda_{\mathrm{normal}}=10$, $\lambda_{\mathrm{mask}}=1.0$, and $\lambda_{\mathrm{Lap}}=0.02$.
In addition, the low-frequency components of rendered normal maps  are extracted using
Gaussian smoothing with a fixed kernel size of $21\times21$ and a standard deviation of $\sigma=5$.
The entire generation and refinement process takes approximately 12 minutes per vehicle.
{A more detailed breakdown of the runtime for each stage is provided in the appendix.}

\noindent \textbf{Datasets.} 
We conduct evaluations on two distinct vehicle  datasets representing synthetic and real-world domains.
SketchFab-Cars is a self-collected dataset consisting of $50$ high-quality 3D vehicle models covering diverse shapes and styles downloaded from SketchFab\footnote{https://sketchfab.com}.
For each model, we  randomly render a view from an azimuth angle from $\{45^\circ, 135^\circ,$ $ 225^\circ, 315^\circ\}$ with elevation of $0^\circ$ as the test input view.
3DRealCar~\cite{du20243drealcar} is  an in-the-wild multi-view vehicle image dataset  containing diverse viewpoints, lighting conditions, and background environments.
Compared to synthetic renderings, images in 3DRealCar exhibit more complex appearance variations, posing greater challenges for both texture synthesis and geometry refinement.
From this dataset, we randomly select $50$ distinct vehicle images to serve as conditional inputs. While SketchFab-Cars enables the evaluation of geometric accuracy, 3DRealCar is restricted to assessing texture fidelity.



\noindent \textbf{Baselines.}   
We evaluate HiFiVe against three categories of baseline methods:
(i) car-specific reconstruction methods, including DreamCar~\cite{du2024dreamcar};
(ii) multi-view diffusion-based methods, including Unique3D~\cite{wu2024unique3d}, InstantMesh~\cite{xu2024instantmesh}, and Hunyuan3D (Paint) 2.1~\cite{hunyuan3d2025hunyuan3d2.1}; and
(iii) native 3D generative models, including Hunyuan3D (Shape) 2.1~\cite{hunyuan3d2025hunyuan3d2.1}, TRELLIS~\cite{xiang2024trellis}, and TRELLIS.2~\cite{xiang2025trellis2}.
All methods are evaluated using their officially released implementations with default inference settings.
Furthermore, we use TRELLIS as the primary benchmark, since our method directly builds upon its generated coarse meshes. This allows for a  controlled comparison to isolate the effect of our texture and geometry refinement stages.

\noindent \textbf{Metrics.}   
We conduct a comprehensive quantitative evaluation with a primary focus on texture quality and geometry enhancement.
For texture evaluation, we report  FID and KID~\cite{binkowski2018demystifying} to measure  visual fidelity, along with two no-reference image quality metrics, MANIQA~\cite{yang2022maniqa} and LIQE~\cite{zhang2023blind}, to assess  perceptual realism.
For geometry evaluation, we report ULIP~\cite{xue2023ulip} and Uni3D~\cite{zhou2023uni3d} scores to evaluate  semantic alignment between the reconstructed shapes and input images, and Chamfer Distance (CD) to measure geometric accuracy.

\begin{figure*}[!h]
\begin{center}
\includegraphics[width=0.9\linewidth]{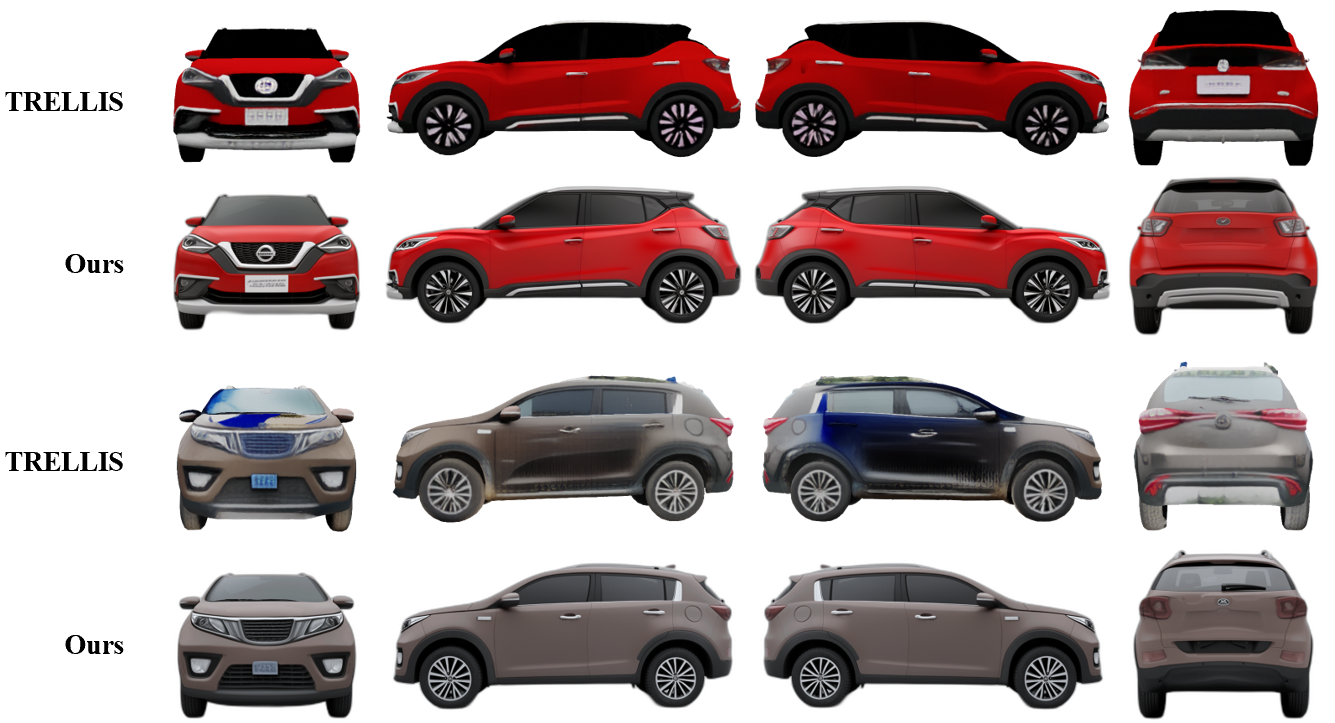}
\end{center}
\vspace{-6mm}
   \caption{ The comparison of texture refinement results with the original textures from TRELLIS coarse mesh.
   }
\label{fig:mv_comp_with_TRELLIS}
\end{figure*}

\begin{table*}[!h]
\centering
\small
\renewcommand{\arraystretch}{0.9}
\setlength{\tabcolsep}{10pt}

\begin{tabular}{c cccc}
\toprule
\multirow{3}{*}{Method} 
& \multicolumn{4}{c}{Metrics} \\

\cmidrule(lr){2-5}

& \multicolumn{2}{c}{Fidelity} 
& \multicolumn{2}{c}{Realism} \\

\cmidrule(lr){2-3} \cmidrule(lr){4-5}

& FID$\downarrow$ 
& KID$\downarrow$ 
& MANIQA$\uparrow$ 
& LIQE$\uparrow$ \\
\midrule

\multicolumn{5}{c}{\textbf{SketchFab-Cars}} \\
\midrule
DreamCar~\cite{du2024dreamcar}        & 210.19 & 0.1271 & 0.4334 & 1.8883 \\
Unique3D~\cite{wu2024unique3d}        & 134.18 & 0.0739 & 0.4936 & 3.7768 \\
InstantMesh~\cite{xu2024instantmesh}     & 113.99 & 0.0503 & 0.4532 & 3.7745 \\
Hunyuan3D 2.1~\cite{hunyuan3d2025hunyuan3d2.1}   & 118.75 & 0.0496 & 0.5371 & 4.6395 \\
TRELLIS.2~\cite{xiang2025trellis2}       & 125.64 & 0.0543 & 0.5169 & 3.8947 \\
\midrule
TRELLIS~\cite{xiang2024trellis}         & 97.48  & 0.0351 & 0.5015 & 4.4924 \\
Ours            & \textbf{95.89} & \textbf{0.0340} & \textbf{0.5666} & \textbf{4.9278} \\
\midrule

\multicolumn{5}{c}{\textbf{3DRealCar}} \\
\midrule
DreamCar~\cite{du2024dreamcar}        & 263.89 & 0.1624 & 0.4477 & 2.5769 \\
Unique3D~\cite{wu2024unique3d}        & 138.11 & 0.0663 & 0.4910 & 4.3232 \\
InstantMesh~\cite{xu2024instantmesh}     & 104.19 & 0.0400 & 0.4566 & 4.2597 \\
Hunyuan3D 2.1~\cite{hunyuan3d2025hunyuan3d2.1}   & 94.01  & 0.0321 & 0.5461 & 4.8817 \\
TRELLIS.2~\cite{xiang2025trellis2}       & 94.02  & 0.0298 & 0.5191 & 4.7182 \\
\midrule
TRELLIS~\cite{xiang2024trellis}         & 90.42  & 0.0284 & 0.4790 & 4.6074 \\
Ours            & \textbf{85.35} & \textbf{0.0272} & \textbf{0.5657} & \textbf{4.9528} \\
\bottomrule
\end{tabular}

\caption{Quantitative comparison of texture quality on both the SketchFab-Cars dataset and 3DRealCar dataset.}
\label{tab:tex_comp}
\end{table*}

\begin{figure*}[!ht]
\begin{center}
\includegraphics[width=\linewidth]{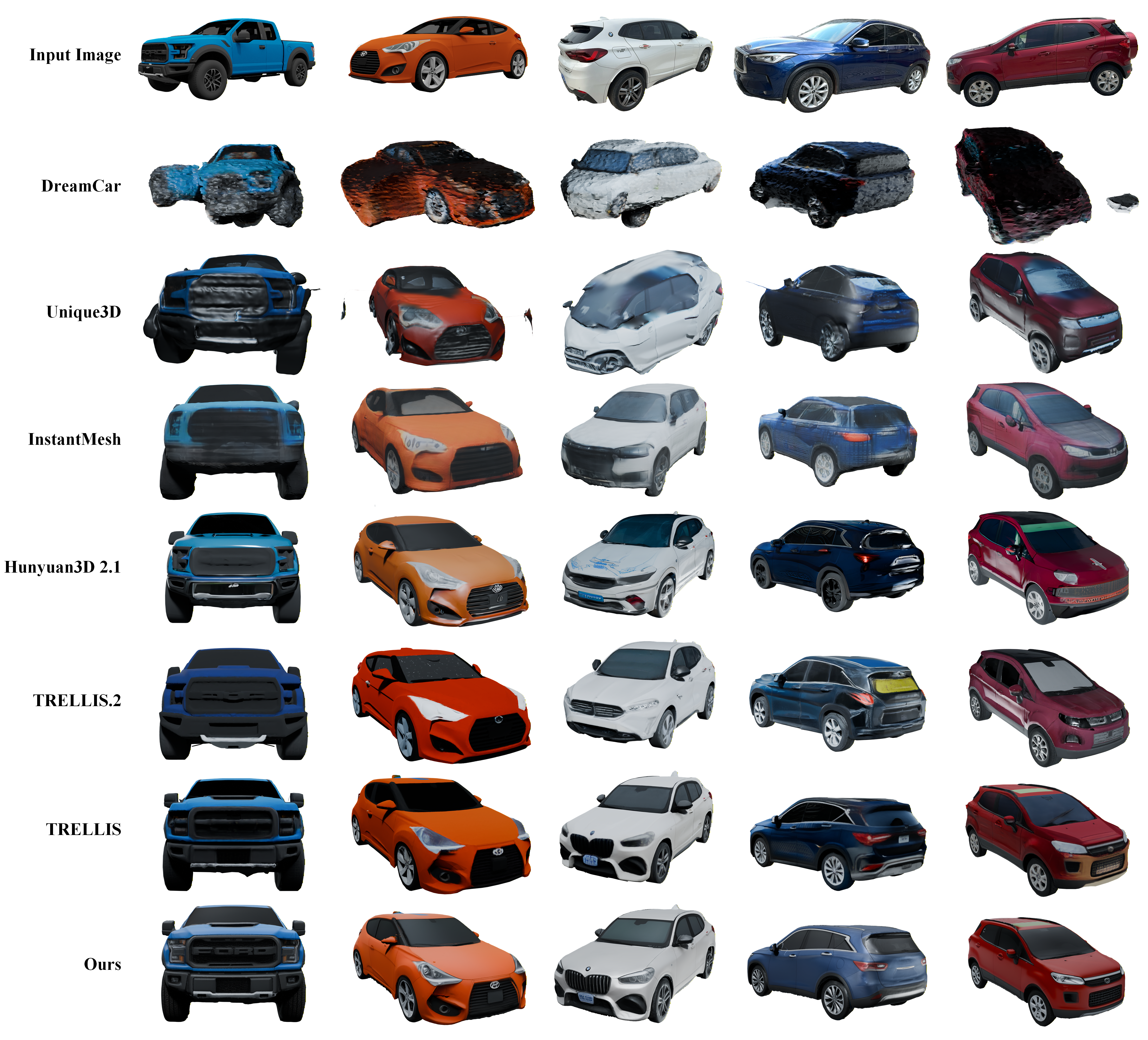}
\end{center}
\vspace{-6mm}
   \caption{  
   Qualitative comparison with baseline methods in texture quality.
   The first two columns show results on the SketchFab-Cars dataset, while the remaining columns correspond to the 3DRealCar dataset.
   }
\label{fig:comp_tex}
\end{figure*}

\subsection{Evaluation of Experimental Results}

We present quantitative and qualitative comparisons with state-of-the-art baselines on both the SketchFab-Cars and 3DRealCar~\cite{du20243drealcar} datasets.

\noindent \textbf{Evaluation on Texture Quality.}
{In Fig.~\ref{fig:mv_comp_with_TRELLIS}, we compare the texture refinement results with the original texture from TRELLIS~\cite{xiang2024trellis} coarse mesh.
Owing to the powerful 2D image prior and our autoregressive refinement pipeline, our method produces clearer multi-view textures with improved cross-view coherence.}
Then we conduct the quantitative and qualitative comparisons on the rendering result from the final optimized mesh.
As shown in Table~\ref{tab:tex_comp},
our method achieves the best performance across all texture metrics on both datasets, indicating superior visual fidelity and perceptual realism compared with SOTA baselines.
Similar conclusion can also be drawn from Fig.~\ref{fig:comp_tex}. 
DreamCar~\cite{du2024dreamcar} suffers from blurred textures and unsmooth geometry surface, leading to the loss of fine details. 
Unique3D~\cite{wu2024unique3d} is poor at handling non-frontal image input, and often produce distorted geometry structures.
InstantMesh~\cite{xu2024instantmesh} and Hunyuan3D-Paint 2.1~\cite{hunyuan3d2025hunyuan3d2.1} jointly generate multiple views through multi-view diffusion, limiting per-view resolution and leading to over-smoothed  textures.
For native 3D generative models, TRELLIS~\cite{xiang2024trellis} and TRELLIS.2~\cite{xiang2025trellis2} typically produce low-resolution textures that lack high-frequency details.
Note that although TRELLIS.2~\cite{xiang2025trellis2} produces better geometry details than TRELLIS~\cite{xiang2024trellis}, it suffers from a noticeable color drift where the generated textures deviate from the input image, leading to reduced visual fidelity.
In contrast, HiFiVe can generate sharper and more realistic textures with strong cross-view consistency. 
Car-specific details such as the headlights, grilles, windows, and body panel boundaries are faithfully preserved, even some brand logos are reconstructed with high fidelity image generation. 


\begin{table*}[]
    \centering
    \small
    \renewcommand{\arraystretch}{0.9}
    \setlength{\tabcolsep}{12pt}
    \begin{tabular}{cccc cc}
        \toprule
        \multirow{2}{*}{Method} & \multicolumn{3}{c}{SketchFab-Cars} & \multicolumn{2}{c}{3DRealCar} 
        \\ 
        \cmidrule(lr){2-4} \cmidrule(lr){5-6}
        & ULIP$\uparrow$ & Uni3D$\uparrow$ & CD$\downarrow$ & ULIP$\uparrow$ & Uni3D$\uparrow$ \\
        \midrule
         DreamCar~\cite{du2024dreamcar} & 0.0859 & 0.1998  & 0.0559 & 0.0391 &  0.1528 \\
         Unique3D~\cite{wu2024unique3d} & 0.1108	& 0.2099 &  0.0551  & 0.1072	& 0.2066  \\
         InstantMesh~\cite{xu2024instantmesh} & 	0.1381 &	0.2423 & 0.0230 &   0.1577 &	0.2630 \\

         Hunyuan3D 2.1~\cite{hunyuan3d2025hunyuan3d2.1} & 0.1604	& 0.3219	& \textbf{0.0103} & 0.1544 &	0.2843  \\
         TRELLIS.2~\cite{xiang2025trellis2} & \textbf{0.1713}	& \textbf{0.3290} &	0.0117  &  0.1423	& 0.2783\\
        \midrule
        TRELLIS~\cite{xiang2024trellis} & 0.1526	& 0.3282	& 0.0116   & 0.1541 &	0.2888 \\
        Ours & 0.1612	& 0.3187	& 0.0119 &  \textbf{0.1621} &	\textbf{0.2904 } \\
        \bottomrule
    \end{tabular}
        \caption{Quantitative comparison of geometry quality on both the SketchFab-Cars dataset and 3DRealCar dataset.}
    \label{tab:geo}
\end{table*}

\begin{figure*}[!ht]

\begin{center}
\includegraphics[width=\linewidth]{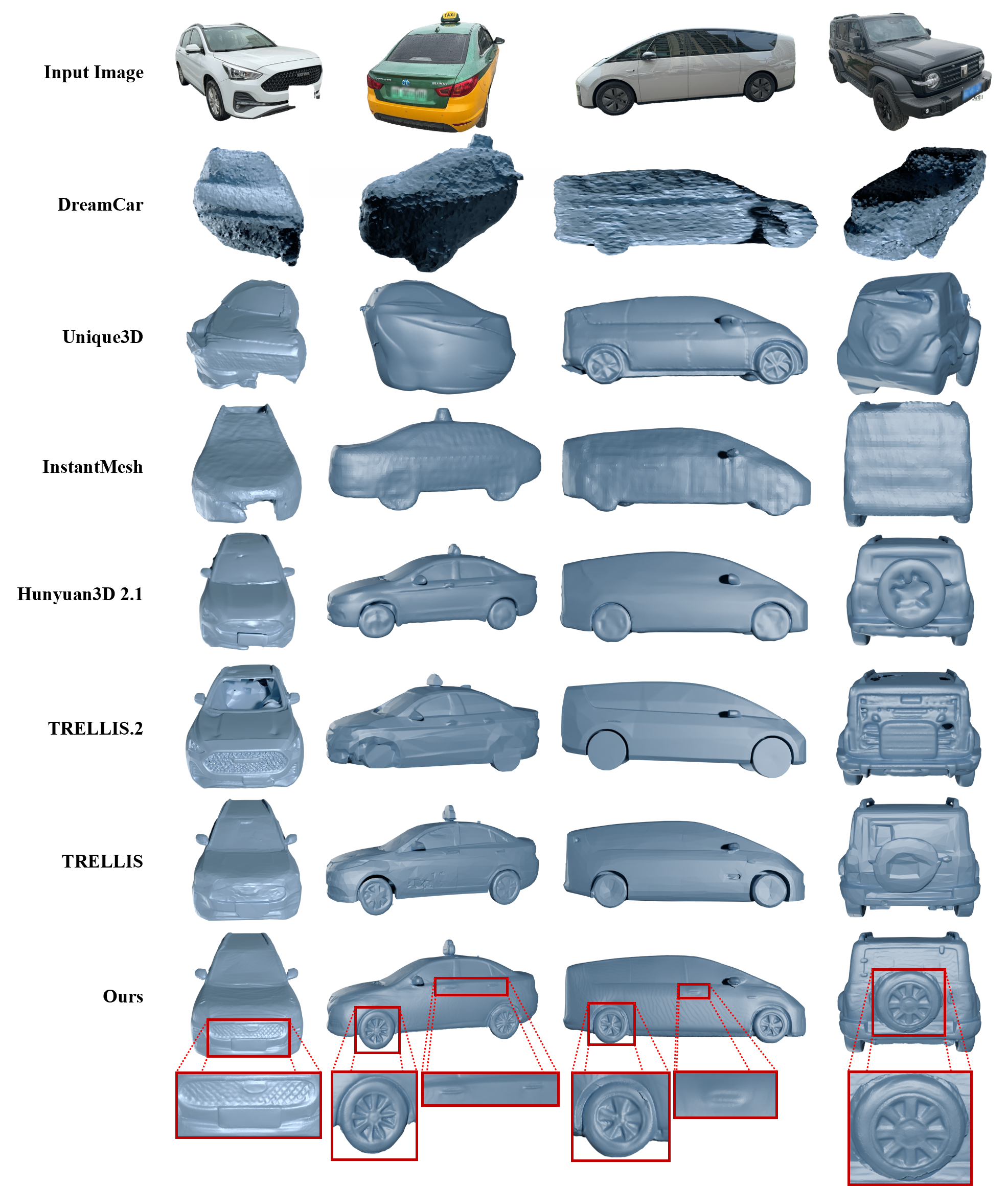}
\end{center}
\vspace{-6mm}
   \caption{Qualitative comparison of geometry reconstruction on the 3DRealCar dataset. 
   Zoomed-in regions highlight representative geometric details recovered by HiFiVe.
}
\label{fig:comp_geo}
\end{figure*}

\noindent \textbf{Evaluation on Geometry Quality.}
To evaluate the geometry refinement improvement, we report the quantitative evaluation of geometric accuracy and semantic alignment in Table~\ref{tab:geo}.
On the SketchFab-Cars dataset, HiFiVe achieves competitive performance compared with TRELLIS series. 
We attribute this to the fact that SketchFab-Cars aligns closely with the training distribution of native 3D generative models, which are typically pre-trained on millions of synthetic 3D assets.  Consequently, the room for geometric improvement provided by our method on this specific dataset is limited.
In contrast, on the more challenging real-world image dataset, \textit{i.e.} 3DRealCar~\cite{du20243drealcar}, HiFiVe outperforms other baseline methods in terms of semantic alignment, indicating the refined geometry is more accurately aligned with the input image.

Also, we present the qualitative comparison on the 3DRealCar~\cite{du20243drealcar} dataset in Fig.~\ref{fig:comp_geo} to further validate the effectiveness of our method. 
As we can see, InstantMesh~\cite{xu2024instantmesh}  suffers from over-smoothed geometry due to the limited resolution of triplane representation. 
Though native 3D generative models~\cite{hunyuan3d2025hunyuan3d2.1, xiang2025trellis2, xiang2024trellis} produce vehicles with richer structural components, they often fail to recover fine-grained surface details. This limitation is inherent, as the reconstruction of detailed geometry is constrained by the resolution and representational capacity of the latent space (e.g., the SLAT \cite{xiang2024trellis} in TRELLIS). In contrast, our optimization-based geometric refinement is not bound by these constraints, and can leverage the state-of-the-art normal prior.
As shown in Fig.~\ref{fig:comp_geo}, the fine details such as wheel spokes, air intake grille, and door handles are much clearer. 
More visualization results are included in~\ref{appendix:additional_qualitative_results}.

\begin{table*}[!ht]
    \centering
    \small
    \renewcommand{\arraystretch}{1.05}
    \setlength{\tabcolsep}{4pt} 
    \begin{tabular}{c|cccc|cccc}
        \toprule
        ID 
        & \makecell{MV\\Diff.}
        & \makecell{1$\rightarrow$1\\Warp}
        & \makecell{$n\rightarrow$1\\Warp}
        & \makecell{Sym.}
        & FID$\downarrow$
        & KID$\downarrow$
        & LIQE$\uparrow$
        & SCS-LPIPS$\downarrow$ \\

        \midrule
        1 (TRELLIS) 
        &  &  &  & 
        & 90.42 & 0.0284  & 4.6074 &  0.0463
        \\
        2 
        & \checkmark &  &  &  
        & 96.83 & 0.0347 & 4.8914 & 0.0413 \\
        3 
        &  & \checkmark &  &  
        & 86.14 & 0.0298 & 4.9428 & 0.0690 \\
        4 
        &  & & \checkmark &  
        & \textbf{84.39} & 0.0283 & 4.9528 & 0.0663 \\
        5 (Ours) 
        &  &  & \checkmark & \checkmark 
        & 85.35 & \textbf{0.0272} & \textbf{4.9528} & \textbf{0.0202} \\
        \bottomrule
    \end{tabular}
    \caption{
    Ablation of  texture generation strategies.
    \footnotesize MV Diff.: multi-view diffusion. Warp: depth-based texture warping. Sym.: symmetry-aware generation trajectory.
    }

    \label{tab:ablation_tex}
\end{table*}

\begin{figure*}[!ht]
\begin{center}
\includegraphics[width=1.0\linewidth]{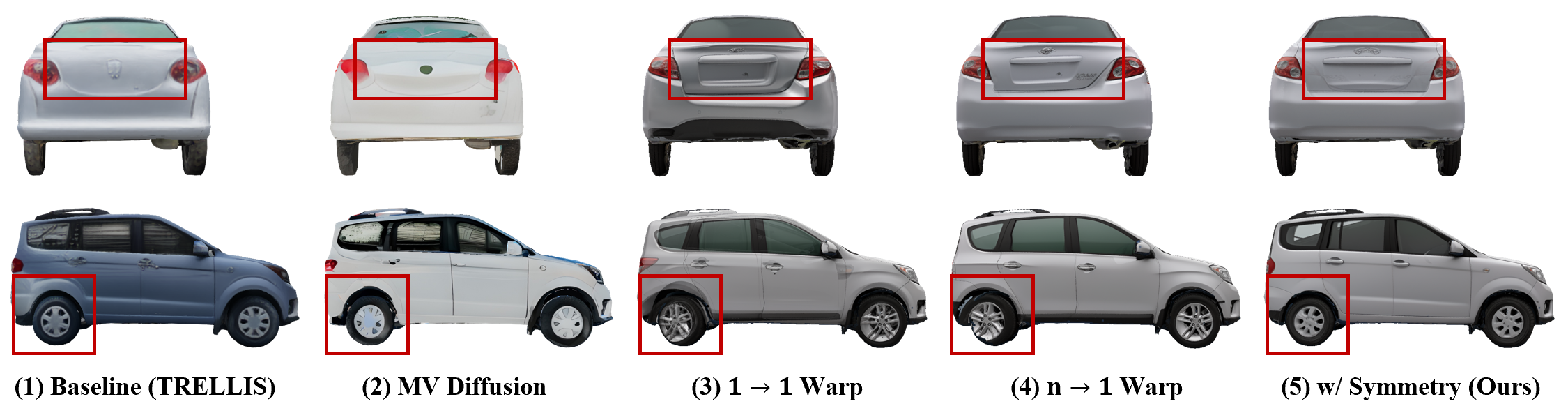}
\end{center}
\vspace{-6mm}
   \caption{  
   Qualitative ablation of  texture generation strategies.
   }
\label{fig:ablation_tex}
\end{figure*}

\begin{table}[!ht]
    \centering
     \renewcommand{\arraystretch}{0.9}
     \label{tab:ablation_recon_view_nums}
    \begin{tabular}{c|c|ccc}
        \toprule
       ID & Method &  ULIP$\uparrow$ & Uni3D$\uparrow$  \\
        \midrule
            1 & Baseline (TRELLIS) &  0.1541 &	0.2888  \\
           2 & Refine w/o FAW &   \textbf{ 0.1617 }&	0.2898   \\
           3 (Ours) & Refine w/ FAW  &   0.1612	& \textbf{0.2926 }   \\
           
        \bottomrule
    \end{tabular}
    \caption{ 
         Ablation of geometry refinement.
    }
    \label{tab:ablation_geo}
\end{table}

\begin{figure}[tbp]
\begin{center}
\includegraphics[width=0.7\linewidth]{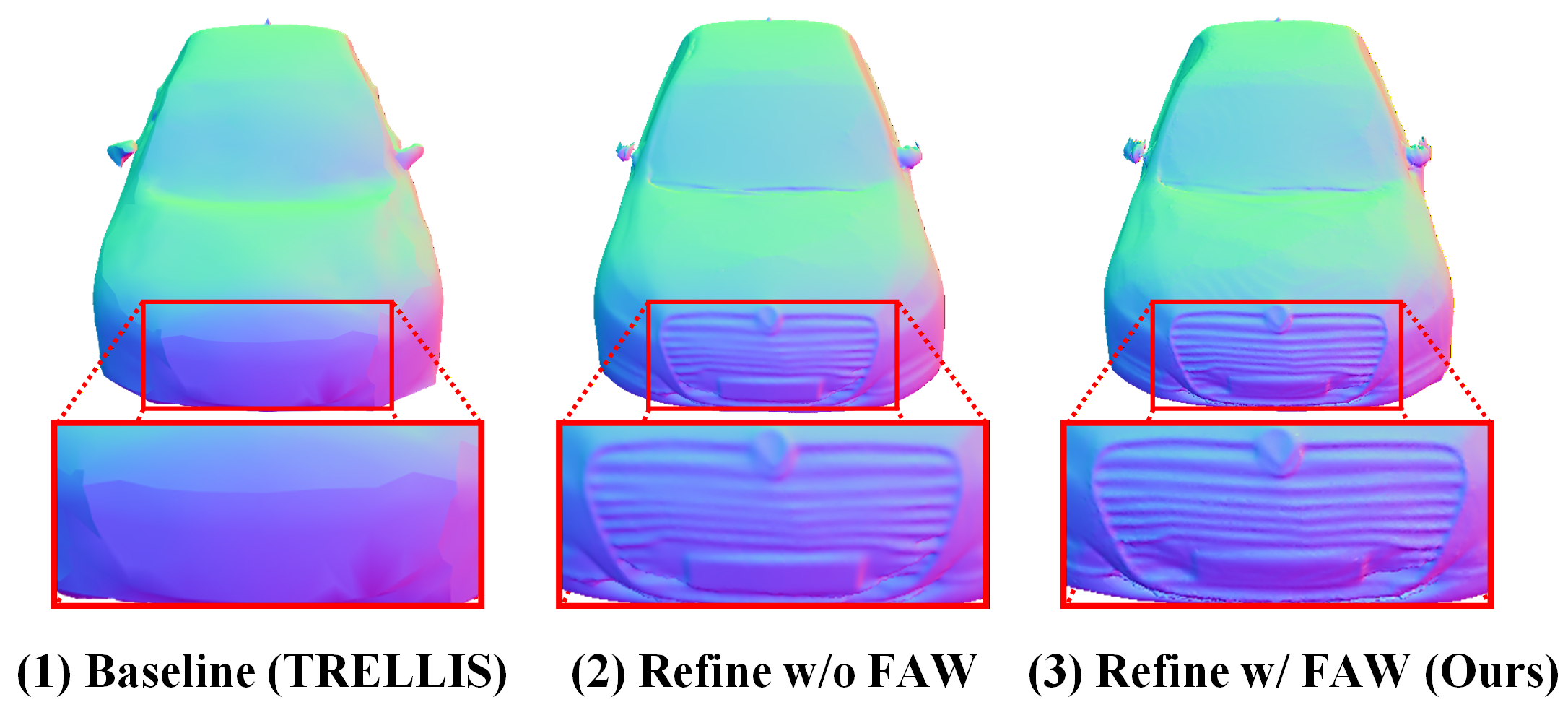}
\end{center}
\vspace{-6mm}
   \caption{ 
  Qualitative ablation of geometry refinement.
    }
\label{fig:ablation_geo}
\end{figure}

 \subsection{Ablation Studies}
We conduct ablation experiments on the 3DRealCar~\cite{du20243drealcar} dataset to validate the contribution of each component in our framework. 

\noindent \textbf{Ablation on Texture Refinement.}
We investigate the effectiveness of  different texture generation strategies based on the same coarse geometry produced by TRELLIS~\cite{xiang2024trellis}. Quantitative and qualitative results are reported in Table~\ref{tab:ablation_tex} and Fig.~\ref{fig:ablation_tex}.
In addition to standard fidelity and perception metrics, we introduce a Symmetry Consistency Score (SCS) to explicitly measure the consistency of textures at symmetrical viewpoints, which is particularly relevant for vehicle generation.
Please refer to~\ref{appendix:scs} for detailed definitions of SCS.

Starting from the TRELLIS baseline (ID 1), the generated textures are globally plausible but exhibit blurred local details (Fig.~\ref{fig:ablation_tex} (1)).
We further evaluate a multi-view diffusion baseline (ID 2) that applies Hunyuan3D-Paint~2.1~\cite{hunyuan3d2025hunyuan3d2.1} to generate textures under the same coarse geometry. 
This method improves perceptual quality but achieves the worst FID and KID scores, as the jointly generated textures tend to be over-smoothed and prone to slight color drift and local inconsistencies (Fig.~\ref{fig:ablation_tex} (2)).
Replacing it with autoregressive generation using single-view warping (ID 3) further enhances texture fidelity.
However, conditioning each target view on a single preceding  view leads to gradual error accumulation along a long chain, resulting in ``ghost effect'' in the fused texture (Fig.~\ref{fig:ablation_tex} (3)).
The introduction of  multi-view texture fusion (ID 4) alleviates this issue by aggregating information from all previously refined views.
However, it does not reduce the effective autoregressive path length, and error accumulation can still propagate along the generative chain (Fig.~\ref{fig:ablation_tex} (4)).
Finally, incorporating the symmetry-aware trajectory (ID 5) yields the most consistent and visually coherent results (Fig.~\ref{fig:ablation_tex} (5)).
In particular, the SCS-LPIPS score decreased significantly, indicating substantially improved texture consistency between symmetric viewpoints.

\noindent \textbf{Ablation on  Geometry Refinement.}
We further analyze the effects of normal-guided geometry refinement and the proposed frequency-adaptive weighting (FAW) strategy. Quantitative and qualitative  results are shown in Table~\ref{tab:ablation_geo} and Fig.~\ref{fig:ablation_geo}.
Starting from the coarse mesh (ID 1), the baseline geometry exhibits an overly smooth surface and lacks fine structural details (Fig.~\ref{fig:ablation_geo} (1)).
Applying normal-guided refinement (ID 2) leads to consistent improvements in both semantic alignment metrics.
Surface normals estimated from refined textures contain rich high-frequency details and can effectively guide the geometry optimization, but the reconstructed surfaces are still relatively blurry (Fig.~\ref{fig:ablation_geo} (2)).
When FAW is further incorporated (ID 3), the refined geometry exhibits a noticeable improvement in high-frequency surface details.
As shown in the zoomed regions of Fig.~\ref{fig:ablation_geo} (3), fine structures such as front grilles become noticeably sharper and more consistent, while the overall surface remains smooth and structurally stable. 

{

\subsection{Discussions}
\noindent \textbf{Discussion on Initial Mesh Quality.} 
HiFiVe assumes a reasonably plausible coarse mesh as input, since it serves as the geometric synchronization prior in our framework. Our method is robust to missing fine details, as they can be compensated by 2D-prior-based texture refinement and normal-guided geometry optimization. 
This is consistent with our qualitative comparisons, in which TRELLIS provides a reasonable coarse structure, while HiFiVe significantly improves texture fidelity and restores sharper local geometric details. However, if the initial mesh contains severe structural errors, missing major parts, or incorrect topology, the quality of the refined mesh may degrade. We leave the refinement from severely degraded initial meshes as future work.


\begin{figure}[htbp]
\begin{center}
\includegraphics[width=1.0\linewidth]{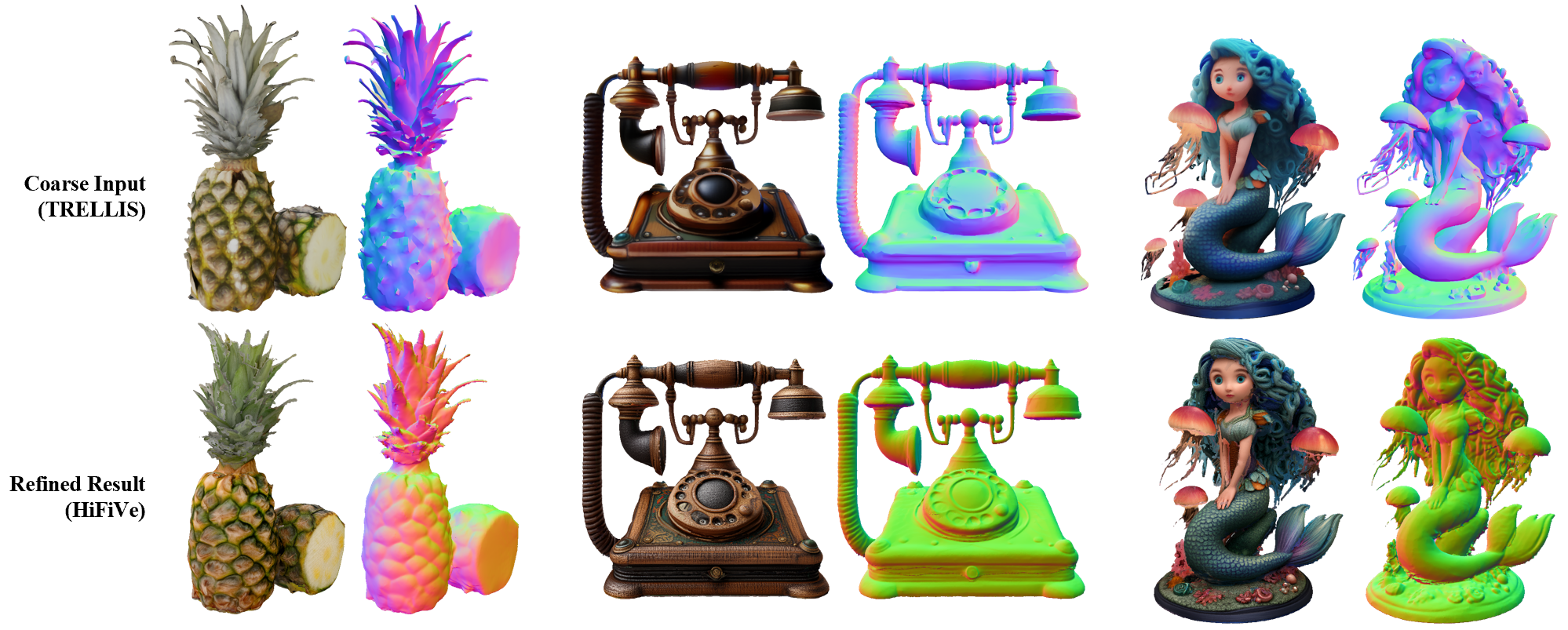}
\end{center}
\vspace{-6mm}
   \caption{ 
   Proof-of-concept extension to non-vehicle categories. 
   To adapt HiFiVe to these examples, we set the anchor-view azimuth to $0^\circ$, use category-specific prompts, and disable the vehicle-specific symmetry shortcut.
   }
\label{fig:extension_other_categories}
\end{figure}

\noindent \textbf{Symmetry Assumption and Extension to Other Categories.}
HiFiVe exploits the approximate bilateral symmetry of common vehicles to shorten the autoregressive generation path. 
However, it may be less suitable for vehicles with asymmetric  appearance or structure, such as one-sided decals, text, damage, or asymmetrical geometry. 
In such cases, the flipped symmetrical views may introduce mirrored textures and fail to preserve side-specific details. 
To address this issue, the symmetry-based shortcut can be disabled, and the corresponding views can instead be refined by the 2D generative model, at the cost of additional inference time.
Beyond vehicles, the main technical components of HiFiVe are not inherently limited to a specific object category. 
Fig.~\ref{fig:extension_other_categories} shows several results on non-vehicle categories, indicating that the proposed framework can be potentially applied to other categories when a reasonable coarse mesh and suitable prompts are provided. 
A systematic evaluation on broader object categories is left as future work.

}

\section{Conclusions}
In this paper, we present HiFiVe, a training-free framework for high-fidelity 3D vehicle refinement.
We introduce a geometry-synchronized autoregressive pipeline to progressively synthesize high-resolution, multi-view textures through depth-based warping and multi-view texture fusion.
This process employs a vehicle-specific symmetric generation trajectory to mitigate long-term error accumulation.
Furthermore, high-frequency surface details can be recovered through normal-guided geometry refinement with frequency-adaptive weighting.
Compared with baseline methods, HiFiVe achieves superior texture fidelity, perceptual realism, and finer geometric quality.
 We believe that our approach will open up new possibilities for scalable and practical 3D car generation, making it a valuable tool for applications in autonomous driving and digital content creation.

\newpage
\appendix


\section{Prompt Design for Texture Refinement}
To ensure stable and photorealistic texture refinement, we adopt carefully designed text prompts for Flux.2~\cite{flux-2-2025} to refine the anchor view and to complete missing regions in warped conditioning images.
{ Unless otherwise specified, all vehicle instances in both datasets use the same prompt templates, without instance-specific prompt tuning.}

\begin{tcolorbox}[
    enhanced,
    colback=gray!8,
    colframe=gray!30,
    arc=2mm,
    boxrule=0.3pt,
    left=6pt,right=6pt,top=6pt,bottom=6pt,
    title=Anchor View Refinement Prompt,
    colbacktitle=gray!50,
    fonttitle=\bfseries,
]
``A photorealistic car under uniform, diffuse studio lighting.

Even illumination with soft light and no harsh highlights.

Matte, evenly lit surfaces with realistic textures and accurate colors.

Pure white background.

No reflections, no specular highlights, no glossy paint,
no HDR effects, no cinematic lighting, no environment reflections.
''
\end{tcolorbox}

\begin{tcolorbox}[
    enhanced,
    colback=gray!8,
    colframe=gray!30,
    arc=2mm,
    boxrule=0.3pt,
    left=6pt,right=6pt,top=6pt,bottom=6pt,
    title=Image Completion Prompt,
    colbacktitle=gray!50,
    fonttitle=\bfseries,
]
``A realistic image of a vehicle.
Complete the missing parts of the vehicle seamlessly, restoring the full car body.

The reconstructed area should match the existing vehicle in color, material, geometry, and lighting.

Preserve the original car design, proportions, and surface details.
Do not alter the visible intact parts of the vehicle.

Maintain consistent reflections, shadows, and highlights.

Photorealistic, high detail, clean edges, realistic car paint and metal materials.

Studio-style neutral background with soft, even lighting.
''
\end{tcolorbox}




\section{Details of Symmetry Consistency Score (SCS)}
\label{appendix:scs}

In addition to standard fidelity and perception metrics, we introduce a Symmetry Consistency Score (SCS) to explicitly evaluate the left--right appearance consistency of generated vehicle textures.
This property is crucial for vehicle generation because approximate bilateral symmetry is an important structural prior.

To compute SCS, we render each reconstructed vehicle from a fixed set of $16$ viewpoints, consisting of $8$ uniformly spaced azimuth angles 
$\theta \in \{0^\circ, 45^\circ, 90^\circ, $ $ 135^\circ, 180^\circ, 225^\circ, 270^\circ, 315^\circ\}$
and two elevation angles of $\phi=0^\circ$ and $\phi=20^\circ$.
For each view $v = (\theta, \phi)$, we define its symmetrical counterpart as $v^{sym} = ((360^\circ - \theta) \bmod 360^\circ, \phi)$.
Views with azimuth angles of $0^\circ$ and $180^\circ$ are excluded because they are self-symmetric and do not provide a meaningful left-right comparison.
The rendered image at $v^{\text{sym}}$ is horizontally flipped to align with the image at $v$, and their perceptual differences are measured using LPIPS~\cite{zhang2018unreasonable}.
The SCS-LPIPS score of a vehicle is obtained by averaging the LPIPS values over all valid symmetric view pairs, and the final dataset score is reported as the mean across all vehicles.
Lower SCS-LPIPS values indicate stronger symmetry consistency.

{
\section{Runtime Analysis}

The runtime comparison is shown in Table~\ref{tab:runtime_comp}. 
We report two variants of our framework: HiFiVe, which uses FLUX.2 [dev] as the default 2D editing model, and HiFiVe-Fast, which replaces FLUX.2 [dev] with the lightweight FLUX.2 [klein] for faster texture refinement. HiFiVe takes an average of 11.6 minutes to refine a vehicle, while HiFiVe-Fast reduces the runtime to 3.67 minutes.
Compared with optimization-intensive methods such as DreamCar~\cite{du2024dreamcar}, MagicBoost~\cite{yang2024magic}, and Elevate3D~\cite{ryu2025elevating}, both HiFiVe and HiFiVe-Fast are more efficient. 
Compared with feed-forward or native 3D generation methods such as InstantMesh~\cite{xu2024instantmesh}, TRELLIS~\cite{xiang2024trellis}, and Hunyuan3D 2.1~\cite{hunyuan3d2025hunyuan3d2.1}, HiFiVe requires longer inference time because it is not designed as a single-pass 3D generator. Instead, it is a training-free refinement framework that improves a coarse mesh through autoregressive high-resolution texture refinement and normal-guided geometry optimization. 
Therefore, its runtime is mainly determined by the adopted 2D generative backbone and the number of autoregressive refinement views.


\begin{table}[t]
\centering
\small
\begin{tabular}{c c}
\hline
Method & Time \\
\hline
DreamCar~\cite{du2024dreamcar} &	134.2 min\\
Unique3D~\cite{wu2024unique3d}	&31.5 s\\
InstantMesh~\cite{xu2024instantmesh}	&9.6 s\\
Hunyuan3D 2.1~\cite{hunyuan3d2025hunyuan3d2.1}&	116.9 s\\
TRELLIS.2~\cite{xiang2025trellis2}&	202.3 s\\
TRELLIS~\cite{xiang2024trellis}&	25.8 s\\
MagicBoost~\cite{yang2024magic}&	20 min\\
Elevate3D~\cite{ryu2025elevating}&	25 min\\

\hline
HiFiVe-Fast & 3.67 min \\
HiFiVe & 11.6 min \\
\hline
\end{tabular}
\caption{Runtime comparison. HiFiVe uses FLUX.2 [dev] as the default 2D editing model, while HiFiVe-Fast replaces it with FLUX.2 [klein] for faster texture refinement.}
\label{tab:runtime_comp}
\end{table}


We further provide a phased runtime breakdown in Table~\ref{tab:runtime_breakdown}. The test was conducted on an A100 80GB GPU. 
For the default HiFiVe, the main computational bottleneck lies in the texture refinement stage, which takes 8.98 minutes and accounts for 77.4\% of the total runtime. 
This is because the autoregressive refinement process invokes the 2D editing backbone multiple times under explicit 3D geometric synchronization. In comparison, coarse mesh generation and geometry refinement only take 0.78 minutes and 1.84 minutes, respectively.
HiFiVe-Fast provides a practical acceleration pathway. By replacing FLUX.2 [dev] with the lightweight FLUX.2 [klein],  the texture refinement time is reduced from 8.98 minutes to 1.05 minutes, resulting in an overall runtime reduction from 11.60 minutes to 3.67 minutes. 
This demonstrates that HiFiVe can directly benefit from faster 2D generative backbones without altering the overall framework.

\begin{table}[t]
\centering
\footnotesize
\setlength{\tabcolsep}{4pt}
\renewcommand{\arraystretch}{1.08}
\begin{tabular}{@{}lccccc@{}}
\toprule
\multirow{2}{*}{\textbf{Stage}} 
& \multicolumn{2}{c}{\textbf{HiFiVe}} 
& \multicolumn{2}{c}{\textbf{HiFiVe-Fast}} 
& \multirow{2}{*}{\textbf{Speedup}} \\
\cmidrule(lr){2-3} \cmidrule(lr){4-5}
& \textbf{Time} & \textbf{Pct.} 
& \textbf{Time} & \textbf{Pct.} 
& \\
\midrule
Coarse mesh generation 
& 0.78 min & 6.7\% 
& 0.78 min & 21.3\% 
& -- \\
Texture refinement 
& 8.98 min & 77.4\% 
& \textbf{1.05 min} & 28.6\% 
& \textbf{8.6$\times$} \\
Geometry refinement 
& 1.84 min & 15.9\% 
& 1.84 min & 50.1\% 
& -- \\
\midrule
\textbf{Total} 
& 11.60 min & 100.0\% 
& \textbf{3.67 min} & 100.0\% 
& \textbf{3.2$\times$} \\
\bottomrule
\end{tabular}
\caption{Runtime breakdown of HiFiVe using different 2D image editing models for texture refinement. HiFiVe uses FLUX.2 [dev], while HiFiVe-Fast replaces it with FLUX.2 [klein]. Percentages are computed relative to the total runtime of each setting.}
\label{tab:runtime_breakdown}
\end{table}

}


{
\section{More Ablations}

\noindent \textbf{Influence of the Confidence Threshold $\tau$.}
The confidence threshold $\tau$ in Eq.~(3) is used to filter unreliable reprojected pixels whose viewing directions are poorly aligned with the surface normals. To analyze its influence, we conduct a small ablation study on 10 samples from the 3DRealCar dataset by varying $\tau$ from $0.05$ to $0.35$. As shown in Table~\ref{tab:tau_ablation}, the overall  performance is relatively stable when $\tau$ varies from $0.05$ to $0.35$, indicating that our texture fusion strategy is not very sensitive to this parameter.  A smaller threshold keeps more reprojected pixels but may include low-confidence observations from oblique views, while a larger threshold removes more unreliable pixels but may also discard useful texture information. We choose $\tau=0.15$ because it achieves the best SCS-LPIPS score, indicating better consistency between symmetrical viewpoints, while maintaining competitive fidelity and perceptual quality.

\begin{table}[t]
\centering
\small
\begin{tabular}{c|ccccc}
\toprule
$\tau$ & FID$\downarrow$ & KID$\downarrow$ & LIQE$\uparrow$ & MANIQA$\uparrow$ & SCS-LPIPS$\downarrow$ \\
\midrule
0.05 & 118.99 & 0.0198 & 4.9549 & 0.5714 & 0.0233 \\
0.15 & 119.69 & 0.0197 & 4.9505 & 0.5693 & \textbf{0.0227} \\
0.25 & 120.87 & 0.0180 & 4.9511 & 0.5700 & 0.0234 \\
0.35 & 122.09 & 0.0195 & 4.9482 & 0.5713 & 0.0246 \\
 \bottomrule
\end{tabular}
\caption{Ablation study on the confidence threshold $\tau$ for multi-view texture fusion.}
\label{tab:tau_ablation}
\end{table}

\noindent \textbf{Influence of the Number of Views.}
To analyze the influence of the number of views in the autoregressive texture refinement stage, we conduct a small ablation study on 10 samples from the 3DRealCar dataset. We compare three settings with 4, 6, and 8 views, and report LIQE~\cite{zhang2023blind}, SCS-LPIPS, and runtime in Table~\ref{tab:view_num_ablation}. LIQE measures the perceptual quality of rendered images, while SCS-LPIPS evaluates the appearance consistency between symmetrical viewpoints.
As shown in Table~\ref{tab:view_num_ablation}, the single-view perceptual quality (measured by LIQE) remains similar under different view numbers. This is reasonable because the local image quality of each refined view is mainly determined by the 2D generative prior. Differently, the number of views has a more significant impact on cross-view consistency. The 8 views achieve the best SCS-LPIPS score, indicating better consistency between symmetrical views. In contrast, fewer views reduce the runtime but may lead to less stable cross-view consistency. Therefore, we choose 8 views as the default setting, as it achieves a better balance between texture consistency and computational cost. 

\begin{table}[t]
\centering
\small
\begin{tabular}{c|ccc}
\toprule
View Num. & LIQE$\uparrow$ & SCS-LPIPS$\downarrow$ & Runtime \\
\midrule
4 & 4.9523 & 0.0401 & 7.4 min \\
6 & 4.9490 & 0.0733 & 9.1 min \\
8 & \textbf{4.9530} & \textbf{0.0232} & 11.6 min \\
 \bottomrule
\end{tabular}
\caption{Ablation study on the number of views.}
\label{tab:view_num_ablation}
\end{table}

\noindent \textbf{Sensitivity to Normal Prediction Errors.}
Since the geometry refinement stage relies on monocular normal estimation, we further analyze the sensitivity of HiFiVe to normal prediction errors. Specifically, we add random perturbations with different noise levels to the estimated normal maps before geometry refinement, and evaluate the resulting meshes on 10 samples from the 3DRealCar dataset, and include TRELLIS~\cite{xiang2024trellis} as the coarse mesh baseline without normal-guided refinement. 
As shown in Table~\ref{tab:normal_noise_ablation}, HiFiVe outperforms TRELLIS when the estimated normals are clean or mildly perturbed, indicating that normal-guided refinement can effectively improve geometric quality. 
When the noise level increases to $\sigma=0.15$, HiFiVe still achieves comparable ULIP and slightly higher Uni3D than TRELLIS, but its performance degrades under stronger perturbation at $\sigma=0.25$.
The qualitative results in Fig.~\ref{fig:normal_noise_ablation} show a similar trend: 
slight perturbations can preserve clear reconstructed details, whereas stronger noise weakens fine structures but does not collapse the overall vehicle geometry.
 This suggests that HiFiVe is reasonably robust to normal prediction errors due to the coarse TRELLIS initialization and multiple refinement constraints.
 




\begin{table}[t]
\centering
\small
\setlength{\tabcolsep}{8pt}
\renewcommand{\arraystretch}{1.05}
\begin{tabular}{lccc}
\toprule
\textbf{Method} & \textbf{Noise Level $\sigma$} & \textbf{ULIP$\uparrow$} & \textbf{Uni3D$\uparrow$} \\
\midrule
TRELLIS & -- & 0.157 & 0.292 \\
\midrule
HiFiVe & 0.00 & 0.164 & \textbf{0.294} \\
HiFiVe & 0.05 & \textbf{0.166} & \textbf{0.294} \\
HiFiVe & 0.15 & 0.157 & 0.293 \\
HiFiVe & 0.25 & 0.147 & 0.290 \\
\bottomrule
\end{tabular}
\caption{
Qualitative sensitivity analysis of geometry refinement to normal prediction errors. TRELLIS denotes the coarse mesh baseline without normal-guided geometry refinement, and $\sigma$ denotes the noise level added to the estimated normal maps before refinement.
}
\label{tab:normal_noise_ablation}
\end{table}

\begin{figure*}
\begin{center}
\includegraphics[width=\linewidth]{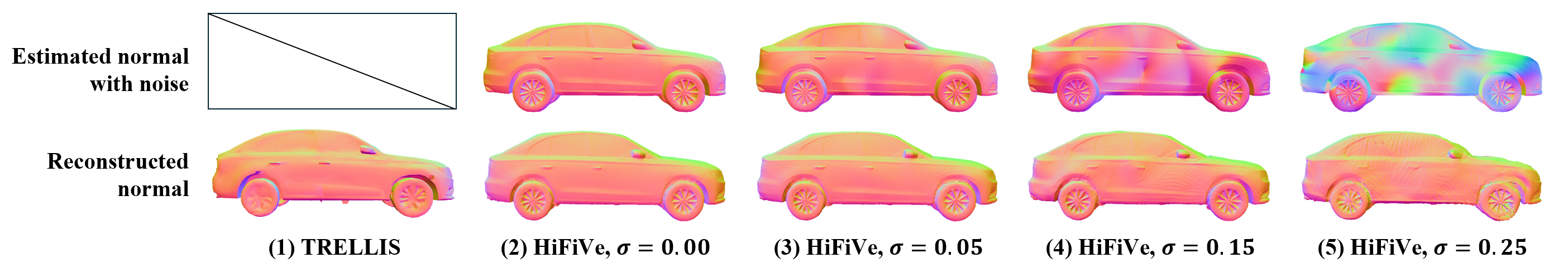}
\end{center}
\vspace{-6mm}
   \caption{ Qualitative sensitivity analysis of geometry refinement to normal prediction errors.}
\label{fig:normal_noise_ablation}
\end{figure*}



}

\section{More Qualitative Results}
\label{appendix:additional_qualitative_results}
\begin{figure*}[!ht]
\begin{center}
\includegraphics[width=1.0\linewidth]{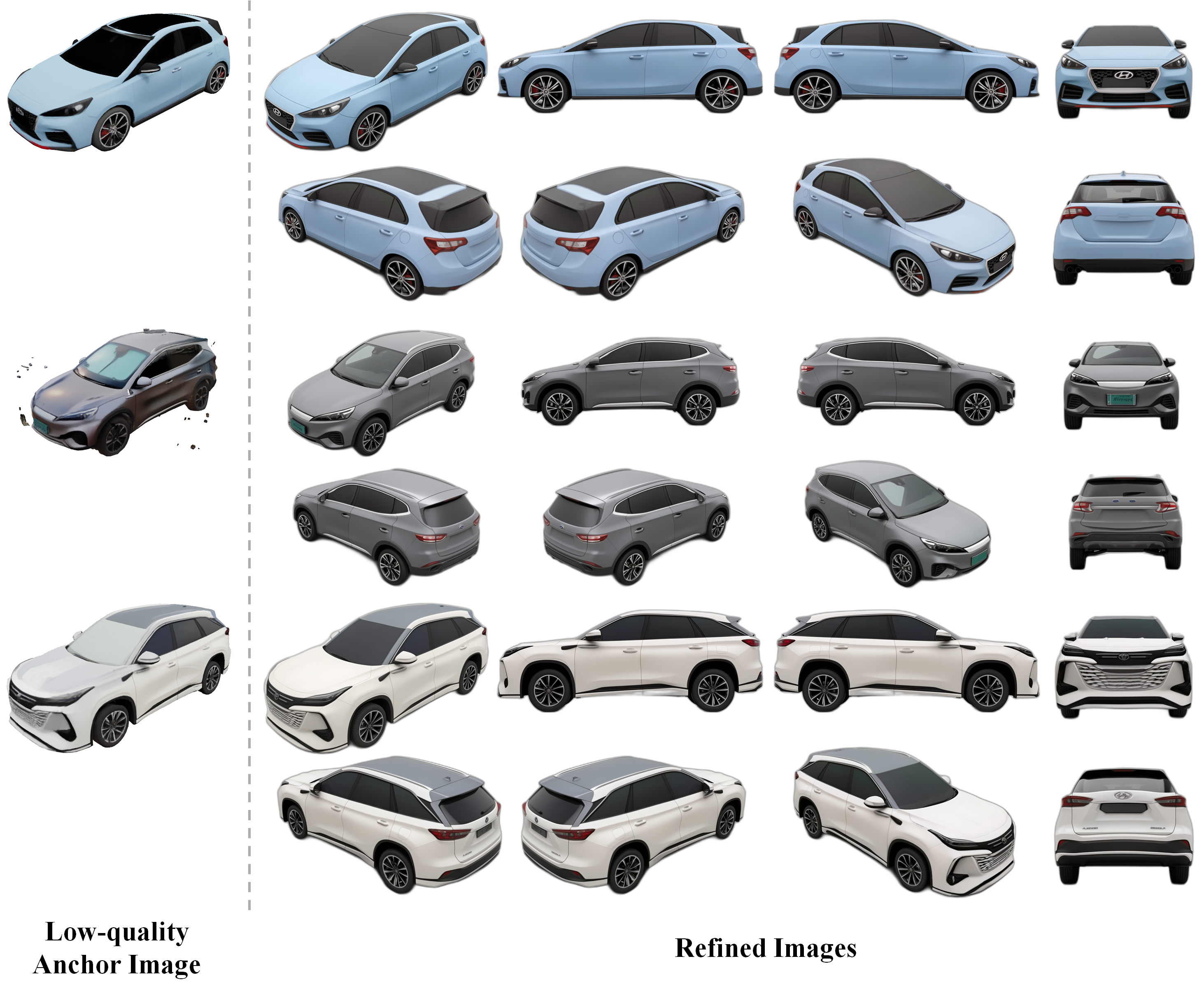}
\end{center}
\vspace{-6mm}
   \caption{
   Intermediate results of autoregressive texture refinement.
   HiFiVe starts with a low-quality anchor image rendered from a coarse mesh and progressively refines the texture across  multiple viewpoints in a geometry-synchronized manner.
   }
\label{fig:trajectory_res}
\end{figure*}

\begin{figure*}[hbp]
\begin{center}
\includegraphics[width=\linewidth]{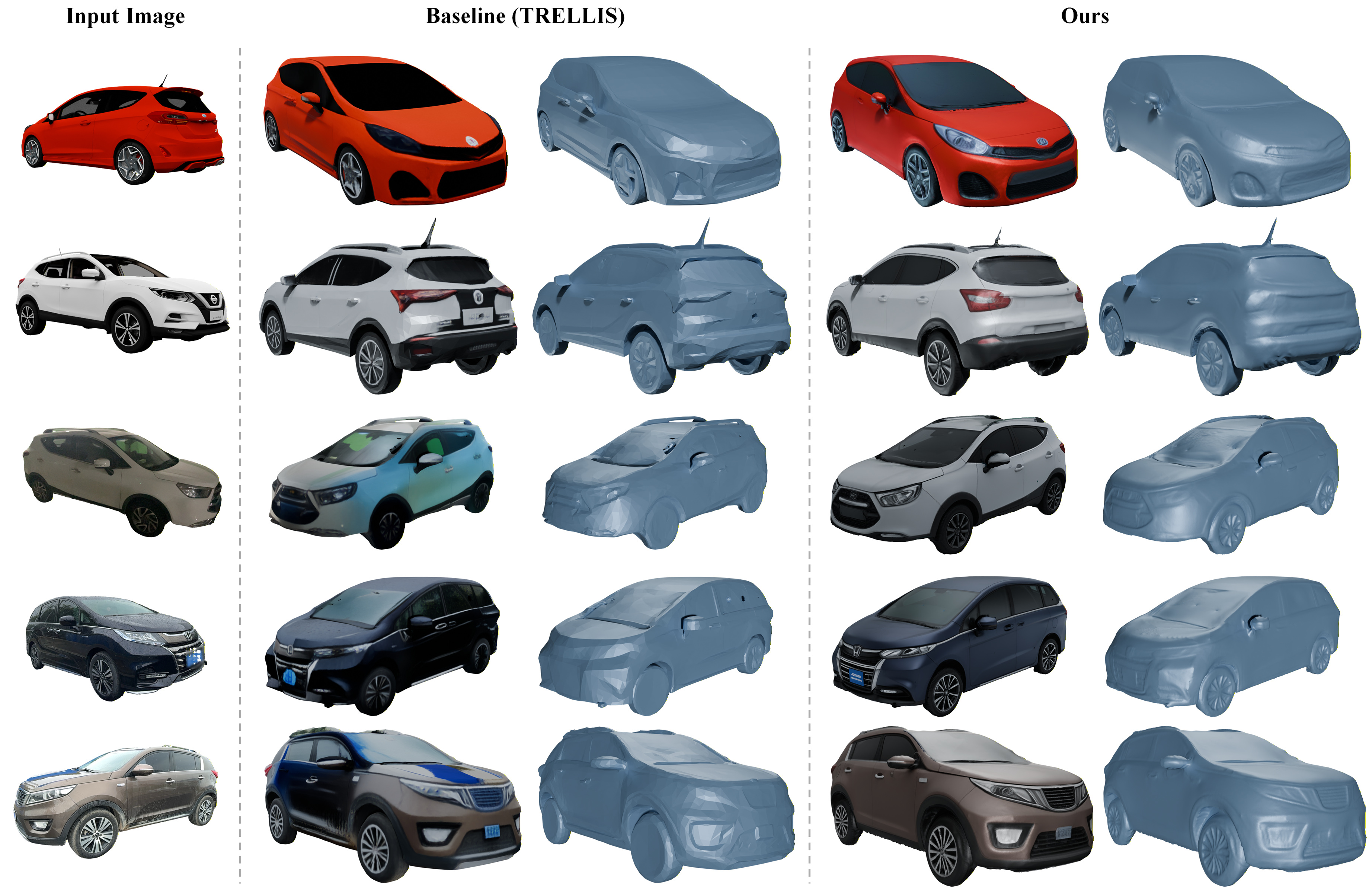}
\end{center}
\vspace{-6mm}
   \caption{ 
   Qualitative comparison between the TRELLIS~\cite{xiang2024trellis} baseline and our method.
   While TRELLIS can generate reasonable, coarse geometry, the resulting shapes and appearances often lack fine details.
   In contrast, our method is able to reconstruct sharper geometric structures and more realistic visual details, benefiting from refined multi-view textures and normal-guided geometry refinement.
   }
\label{fig:vsTRELLIS_only}
\end{figure*}

\begin{figure*}[hbp]
\begin{center}
\includegraphics[width=\linewidth]{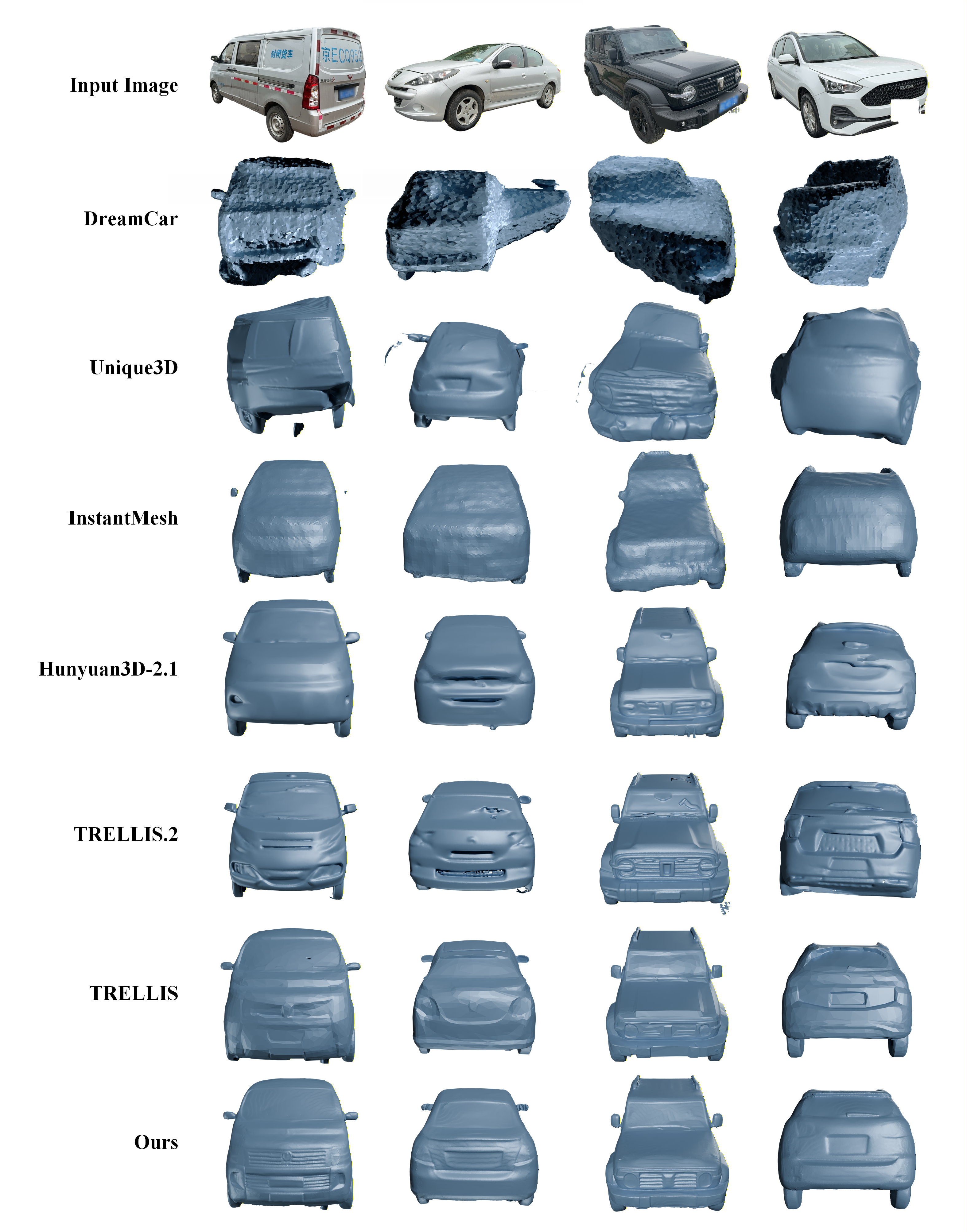}
\end{center}
\vspace{-6mm}
   \caption{ 
   Additional qualitative comparisons of geometry reconstruction results.
   }
\label{fig:geo_comp_appendix}
\end{figure*}

\begin{figure*}[!ht]
\begin{center}
\includegraphics[width=\linewidth]{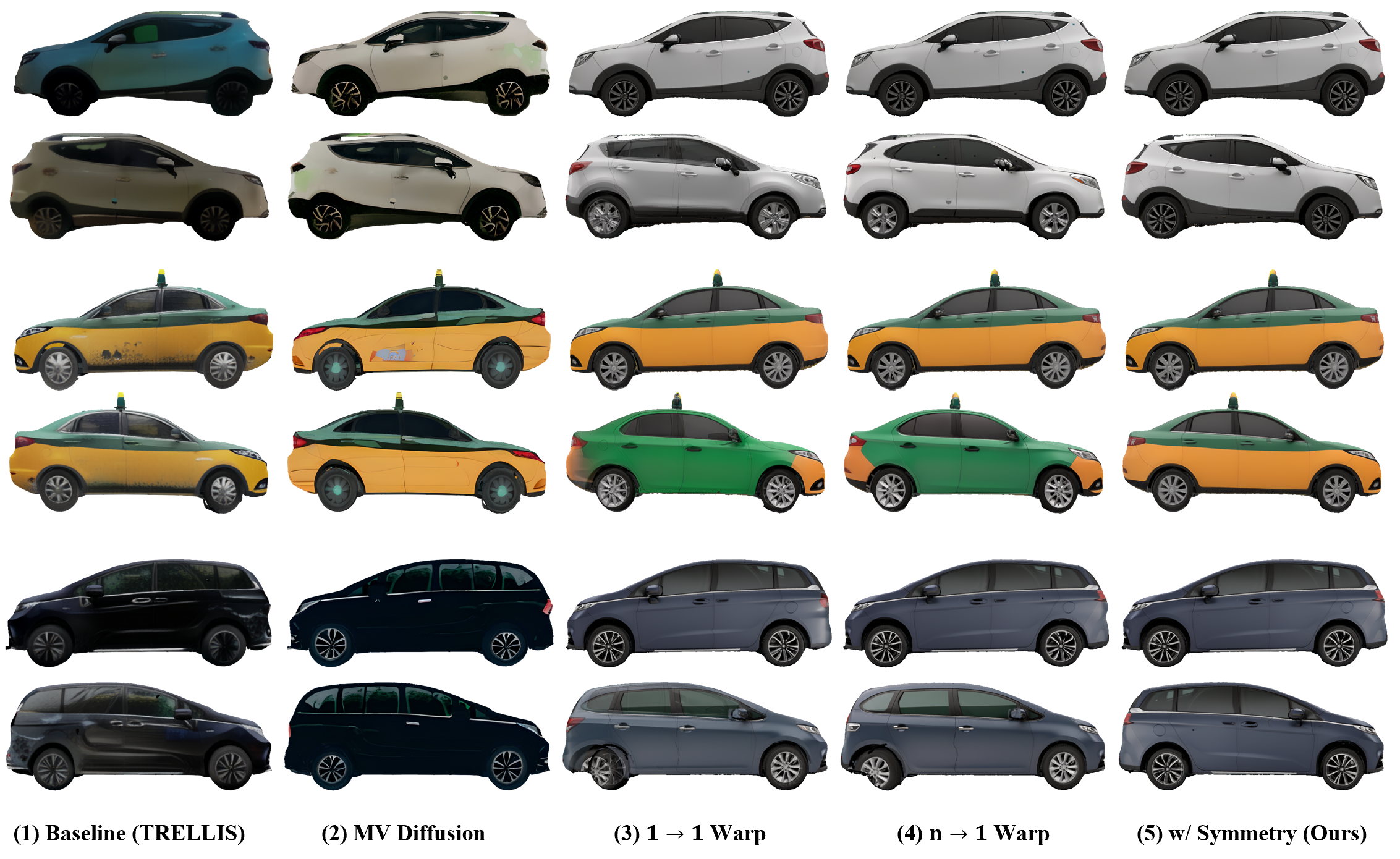}
\end{center}
\vspace{-6mm}
   \caption{ 
   Additional qualitative ablation results of texture refinement strategies.
   Methods without explicit symmetry modeling often produce inconsistent appearances in mirrored views.
   In contrast, our method maintains appearance consistency under symmetrical viewpoints, demonstrating the effectiveness of incorporating vehicle-specific symmetry priors.
   }
\label{fig:ablation_tex_appendix}
\end{figure*}

Additional qualitative results are provided in Fig.~\ref{fig:trajectory_res} - Fig.~\ref{fig:ablation_tex_appendix}.

\clearpage
\bibliographystyle{elsarticle-num} 
\bibliography{ref}








\end{document}